\documentclass{article}
\usepackage{spconf,amsmath,graphicx}

\usepackage{subfigure}
\usepackage{url}
\usepackage{array}
\usepackage{color}
\usepackage{dirtytalk}

\newcommand\blfootnote[1]{%
  \begingroup
  \renewcommand\thefootnote{}\footnote{#1}%
  \addtocounter{footnote}{-1}%
  \endgroup
}

\title{From Video Game to Real Robot: The Transfer between Action Spaces}
\name{Janne Karttunen*\textsuperscript{1, 2} , Anssi Kanervisto*\textsuperscript{2}, Ville Kyrki\textsuperscript{3}, Ville Hautam{\"a}ki\textsuperscript{2} \thanks{* Equal contribution. This research was partially funded by the Academy of Finland (grant \#313970) and Finnish Scientific Advisory Board for Defence (MATINE) project \#2500M-0106. We gratefully acknowledge the support of NVIDIA Corporation with the donation of the Titan Xp and V GPUs used for this research.}}
\address{\textsuperscript{1} Karelics Oy, Joensuu, Finland\\
\textsuperscript{2} School of Computing, University of Eastern Finland, Joensuu, Finland\\
\textsuperscript{3} School of Electrical Engineering, Aalto University, Espoo, Finland \\
\normalsize{janne.a.karttunen@gmail.com, \{anssk, villeh\}@uef.fi, ville.kyrki@aalto.fi }}

\begin{document}
%
\maketitle

\begin{abstract}
Deep reinforcement learning has proven to be successful for learning tasks in simulated environments, but applying same techniques for robots in real-world domain is more challenging, as they require hours of training. To address this, transfer learning can be used to train the policy first in a simulated environment and then transfer it to physical agent. As the simulation never matches reality perfectly, the physics, visuals and action spaces by necessity differ between these environments to some degree. In this work, we study how general video games can be directly used instead of fine-tuned simulations for the sim-to-real transfer. Especially, we study how the agent can learn the new action space autonomously, when the game actions do not match the robot actions. Our results show that the different action space can be learned by re-training only part of neural network and we obtain above $90\%$ mean success rate in simulation and robot experiments.
\end{abstract}


\section{Introduction}
\blfootnote{Copyright 2020 IEEE. Published in the IEEE 2020 International Conference on Acoustics, Speech, and Signal Processing (ICASSP 2020), scheduled for 4-9 May, 2020, in Barcelona, Spain. Personal use of this material is permitted. However, permission to reprint/republish this material for advertising or promotional purposes or for creating new collective works for resale or redistribution to servers or lists, or to reuse any copyrighted component of this work in other works, must be obtained from the IEEE. Contact: Manager, Copyrights and Permissions / IEEE Service Center / 445 Hoes Lane / P.O. Box 1331 / Piscataway, NJ 08855-1331, USA. Telephone: + Intl. 908-562-3966.}
When it comes to training robots to solve a given task with reinforcement learning, one feasible way to do so is by training the policy in a simulation and then using transfer learning \cite{taylor2009transfer} to learn the final policy on the real-world robot; a \textit{simulation-to-real} or \textit{virtual-to-real} transfer \cite{domainRandomization}. This way we are not hindered by expensive and slow robotics experiments. However, simulating real world accurately is hard if not impossible, and thus data obtained from simulation may not be directly applicable to real-world robot, a problem termed \textit{reality gap} \cite{domainRandomization}. To address this, one can try to create as realistic simulation as possible, which requires vast amount of time and is costly.

\begin{figure}[t!]
\centerline{\includegraphics[width=1.0\columnwidth]{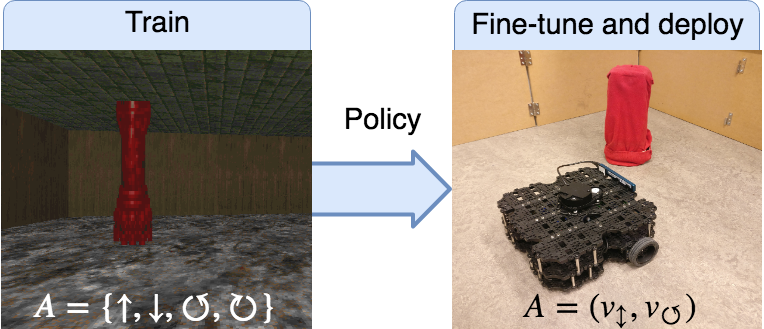}}
\caption{A policy was trained in a video game with an action space consisting of four discrete actions, and then transferred to a robot with a different action space with small amount of training on the robot.}
\label{fig:intro}
\end{figure}

Video games can act as one such simulation: They can be ran fast, are readily available and are shown to be useful in control and reinforcement learning research \cite{mnih2015human,sc2}. Software packages such as ViZDoom \cite{vizdoom} are designed for reinforcement learning. When transferring the trained policies from video games to real-world robots, methods such as domain randomization \cite{domainRandomization} can be used to narrow the reality gap between visual appearances (observations) of the two worlds. However, as video games are not designed for robotics simulations, they lack the options to tune dynamics to match the real world. This leads to a mismatch between available actions between these two worlds. Differences in action dynamics, such as different rotation speeds, could be manually fine-tuned away, but in the case of completely removed actions this is not possible, e.g. when robot is not able to turn left while original simulation allowed this.

In this work we train a \textit{deep reinforcement learning} (DRL) agent which we adapt to a different action space with as little additional training as possible. We demonstrate effectiveness of this method by transferring agent from a crude simulation (video game \say{Doom}, 1993) to a real robot, where the task is the same, environment shares visual similarities, but the action space differs. We conduct experiments with semantically similar action space where the agent can execute similar actions as previously via new action space. We also experiment by removing possible actions from target action-space, effectively hindering agent's capabilities. 

\section{Action Space Transfer in Reinforcement Learning}

\begin{figure}[t]
\begin{center}
\centerline{\includegraphics[scale=0.16]{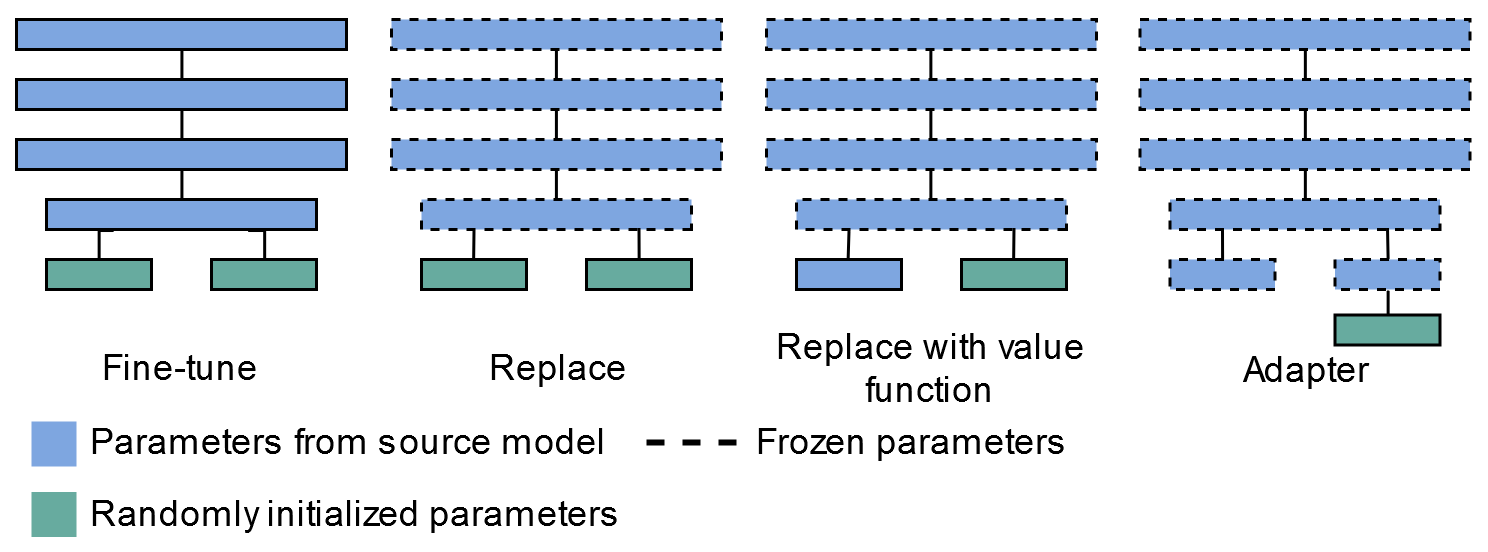}}
\caption{Overview of the different training methods for moving pre-trained neural networks to new action spaces. Boxes represent different layers of a network with left head being value estimation and right being policy or state-action values.}
\label{fig:method}
\end{center}
\vspace{-.5cm}
\end{figure}

Our work is closely related to experiments conducted by Rusu et al. 2016 \cite{rusu2016progressive}, but here we focus on transfer between action-spaces and not tasks.  Video games are a popular benchmarks in reinforcement learning scheme \cite{bellemare13arcade, vizdoom, synnaeve2016torchcraft, kanervisto2018torille}, and video game engines have been used for robotics experiments \cite{shah2018airsim, openai2018dext, HolodeckPCCL}. Our work differs by using a video game as a simulator for robotics experiments successfully, despite the game was not designed for such purpose. We use methods from previous work on simulation-to-real transfer to overcome the visual reality gap \cite{openai2018dext, domainRandomization, Antonova2017ReinforcementLF}, while our contribution lies in bridging the reality gap in action spaces.   

The action space transfer can be done with neural networks by replacing the final (output) layer to fit the new action space. If randomly initialized, this final layer requires some training in the target domain to produce useful actions. At this point we have multiple choices how this training should be done. We opted for four similar and simple methods for our study (see Figure \ref{fig:method}). These methods are similar to baseline methods in \cite{rusu2016progressive}, but here we transfer between action spaces rather than tasks.

{\bf Fine-tuning} Target model uses source model's parameters as initial values, and begins training from there, fine-tuning the previously learned parameters \cite{taylor2009transfer}. This is known to be prone to {\em catastrophic forgetting} \cite{goodfellow2013empirical}, where neural network \say{forgets} the good parameters learned in the previous environment, and thus may not perform as well as expected.

{\bf Replace} We can avoid catastrophic forgetting by not updating some of the neural network parameters at all (\say{freezing}). We freeze all layers except the output layers, since we assume similar dynamics and visual appearance from the two environments, allowing the re-use of features of penultimate layer.

{\bf Replace with pre-trained value function} Value of a state depends on the policy, which depends on the action-space. However, with our assumption of same task and similar environment dynamics between environments, the learned value function could serve as a good initial point in the target environment. We do not freeze the value layer to allow it to adapt to the new policy.

{\bf Adapter} Instead of updating parameters of the source network, we keep them all fixed and learn a mapping from source actions to new actions, essentially learning which action in source environment matches an action in the target environment. Similar method has been used successfully with policy transfer from one domain to another \cite{fernandez2006policy}. We implement this by adding a fully-connected layer which maps old actions to new actions.

\section{Experiments}
\label{sec:experiments}

\subsection{Experimental setup}

Agent's task is to navigate to a red goal pillar in a simple room without obstacles, using visual input as the observation (see Figure \ref{fig:intro}). It starts each episode from the center of the room, facing to a random direction and receives positive reward $1$ for reaching the goal and negative reward $-1$ if episode times out after $1000$ environment steps. Agent chooses an action every $10$ environment steps (frameskip) and receives a color image of size $80 \times 60$ to decide the action. The image is the green channel of a RGB image to highlight the goal.

We use two RL learning algorithms for our experiments: {\em deep Q-learning} (DQN) \cite{mnih2015human} and {\em proximal policy optimization} (PPO) \cite{ppo}. DQN is selected as it is known to be sample efficient, thanks to its off-policy learning and replay memory. Experiments with PPO are included for its applicability to continuous action spaces and for its closer connection to optimizing policy directly. With DQN, we use double Q-learning \cite{van2016deep} and dueling architecture \cite{wang2015dueling} to obtain the state-value function. We use implementations from stable-baselines \cite{stable-baselines}. Both learning algorithms use network described in Mnih et. al. 2015 \cite{mnih2015human}.

To find suitable exploration strategy, we performed hyperparameter search for the exploration parameters during action space transfer with \textit{replace} method. For DQN's $\epsilon$-greedy policy we anneal chance of a random action from $1.0$ to $0.02$ over first $7500$ agent steps (searched over interval $[500, 25000]$). For PPO we tested entropy weight coefficients in interval $[10^{-6}, 1]$ and selected $10^{-3}$ for further experiments. Code and video of the results are available in GitHub \footnote{\url{https://github.com/jannkar/doom_actionspace}}. 

\subsubsection{Source environment}
For the source environment, we use ViZDoom \cite{vizdoom} platform, which is based on the Doom (1993) video game. The agent's action space consists of four different actions; move forward, move backward, turn left and turn right. 

We apply domain randomization to ensure that policy can be transferred to a robot \cite{domainRandomization}. We randomize textures of the walls, ceilings and floors ($68$ different textures), agent's field of view ($50$ to $120$ degrees horizontally), height of the player and head-bobbing strength. We also add small amount of white- and Gaussian noise on the image, and finally apply random amount of gamma-correction ($0.6$ to $1.5$).

\subsubsection{Target environment}
We start experiments by transferring the agent between two Doom environments. In these simulation-to-simulation experiments (\say{\textit{sim-to-sim}}) the environment uses unseen textures and action-space to the agent. For PPO experiments the new action space consists of two continuous values, one for forward/backward velocity and another for rotation speed. For DQN we define discretized action space of $24$ actions, each action being some combination of forward/backward speed and left/right turning, similar to the continuous action space.

The results of best method in sim-to-sim experiments are then verified with real-world experiments by transferring agent to a Turtlebot 3 Waffle robot with a RGB camera. The task is same with a similar environment, with major difference being the lack of roof (Figure \ref{fig:intro}) and larger number of environment steps per one action ($15$ versus original $10$).

\subsection{Source models}

We trained three separate source models with both DQN and PPO in Doom environment, by repeating the training runs. All the following experiments were repeated over these source models, since the source model parameters can affect on final performance of the transfer. All three source models of both algorithms learned to solve the task. The three DQN source models reached a $90-95\%$ testing success rate and mean episode length of $18.57-27.54$ steps, while PPO reached $98-100\%$ success rate with mean of $10.45-14.01$ steps per episode.

\subsection{Sim-to-sim experiments}

\renewcommand{\arraystretch}{1.1}
\begin{table*}[t]
\caption
{
    Mean and standard deviation of final success rates from transferring three different source models with different methods. Each result is based on average performance over last $10\%$ training episodes and averaged over five repetitions. Experiments with average success rate above $90\%$ are highlighted. 
}

\newcolumntype{M}{ m{1.55cm} }
\label{table:simtosim_success}
\begin{center}
\small
\begin{tabular}{MMMMMMM}
                      & \multicolumn{6}{c}{Learning algorithm and source model}                                                                                            \\ \cline{2-7} 
                      Method & \multicolumn{3}{c}{DQN}                                               & \multicolumn{3}{c}{PPO}                                               \\ \hline
       & \multicolumn{1}{c}{1} & \multicolumn{1}{c}{2} & \multicolumn{1}{c}{3} & \multicolumn{1}{c}{1} & \multicolumn{1}{c}{2} & \multicolumn{1}{c}{3} \\ \cline{2-7} 
Fine-tune             & 58.9 $\pm$ 28.7           & \textbf{100 $\pm$ 0.0}             & 54.1 $\pm$ 26.7           & 46.8 $\pm$ 28.4           & 31.7 $\pm$ 14.6           & 52.7 $\pm$ 21.2           \\
Replace               & \textbf{99.9 $\pm$ 0.5}            & \textbf{100 $\pm$ 0.0}             & \textbf{99.2 $\pm$ 3.0}            & \textbf{95.2 $\pm$ 6.7 }           & \textbf{98.0 $\pm$ 2.3}            & \textbf{95.9 $\pm$ 5.4}            \\
Replace  w/ value func. & 60.1 $\pm$ 33.6           & \textbf{95.7 $\pm$ 6.5}            & 86.8 $\pm$ 9.5            & \textbf{98.0 $\pm$ 2.3}            & \textbf{96.1 $\pm$ 3.5}            & \textbf{98.0 $\pm$ 1.3}            \\
Adapter               & 66.3 $\pm$ 31.4           & \textbf{100 $\pm$ 0.0}             & 88.9 $\pm$ 19.0           & 29.7 $\pm$ 8.7            & 30.0 $\pm$ 7.4            & 31.6 $\pm$ 7.8  \\ 
\hline 
Scratch               & \multicolumn{3}{c}{\textbf{99.5 $\pm$ 1.1}}                    & \multicolumn{3}{c}{56.0 $\pm$ 25.4}

\end{tabular}
\end{center}
\end{table*}
\renewcommand{\arraystretch}{1.0}

\begin{figure}[t!]
\centerline{\includegraphics[width=1.0\columnwidth]{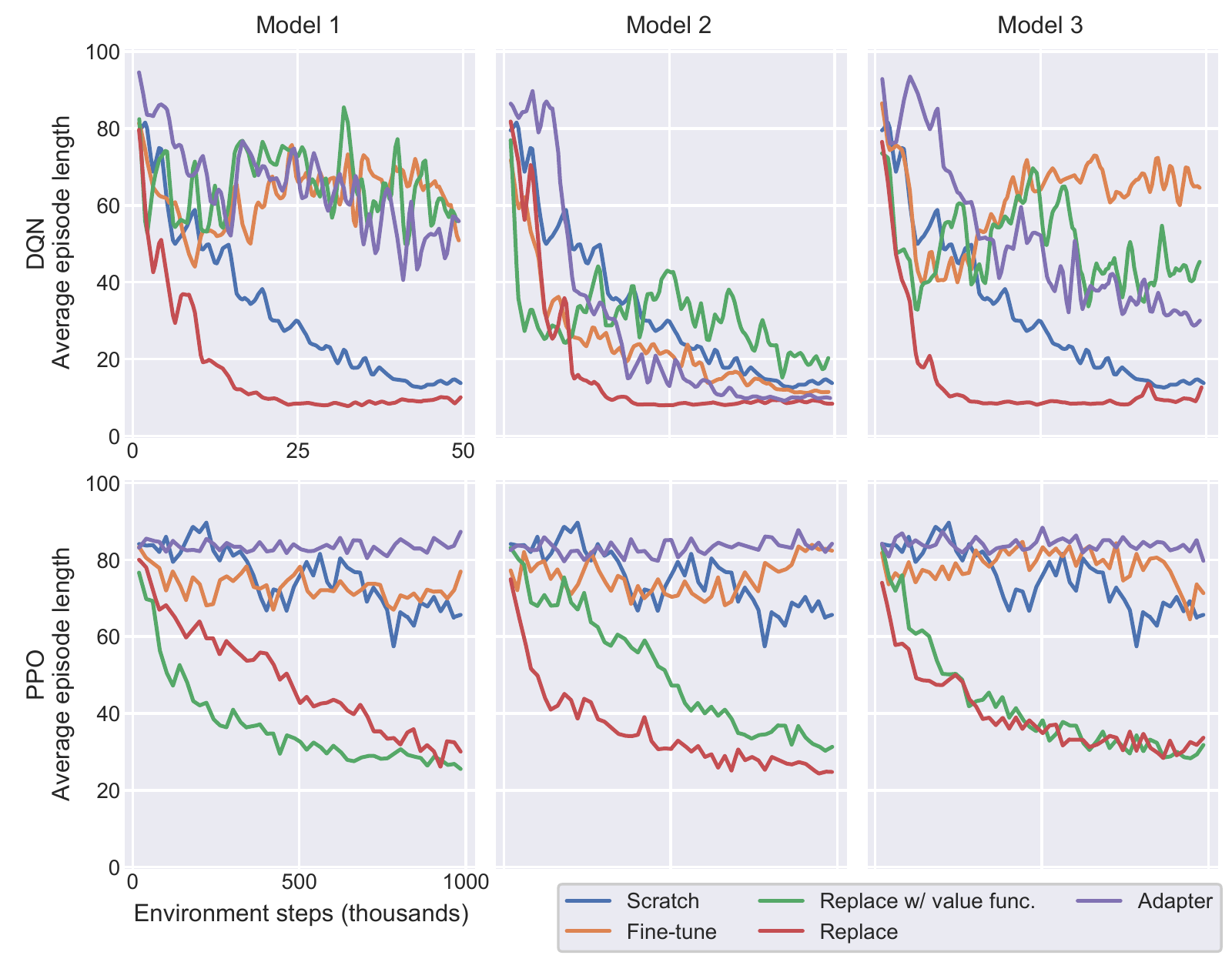}}
\caption{Results of transferring three source models to a new action space with different transfer methods, source models (columns) and learning algorithms. Lower is better. Each line is an average over five repetitions. We omitted variance for visual clarity.
Rows share same baseline result (no transfer learning, "Scratch").}
\label{fig:sim2sim_results}
\end{figure}

Freezing most of the network (\textit{replace} method) performs most reliably with DQN, reaching $99.2-100\%$ success rate (Table \ref{table:simtosim_success}) with the final policy. The task was solved efficiently already at around $15,000-20,000$ steps. Compared to learning the task from scratch, which took approximately $30,000-40,000$ steps for efficient policy, \textit{replace} method reduces the learning time to almost half. Other methods, \textit{adapter}, \textit{fine-tune} and \textit{replace with value function} performed even worse than training from scratch (Figure \ref{fig:sim2sim_results}).
Interestingly, large variance is detected between source model: For source model 2 all tested methods worked well but for models 1 and 3 only \textit{replace} method gave stable performance.

PPO is not as sample efficient algorithm as DQN, and as such it did not learn the task reliably before $1,000,000$ steps. The \textit{replace} method again resulted in robust transfer between action spaces and the performance even improved slightly on average when the value function was loaded with it. However, the task was not solved as efficiently as with the DQN, when considering the episode length. 

In light of these results, \textit{replace} method has the stablest results among tested methods. Interestingly, Rusu et. al. (2016) \cite{rusu2016progressive} found this method least effective in task-transfer scenario, while our results find it most promising in action-space transfer.

\subsection{Robot experiments}

Finally, we validate sim-to-sim experiments on a Turtlebot robot. Based on the previous results, we chose DQN algorithm with the \textit{replace} method for these experiments. We selected DQN model $3$ as the source model, due to its fastest learning in the \textit{replace} method experiment. The agent was trained for $20,000$ steps or until the agent's performance did not increase. Turtlebot takes approximately two actions per second, which translated to $4-5$ hours of wall-clock time per one experiment, including the time to reset the episode.

\begin{figure}[t]
  \centering
  \centerline{\includegraphics[width=1.0\columnwidth]{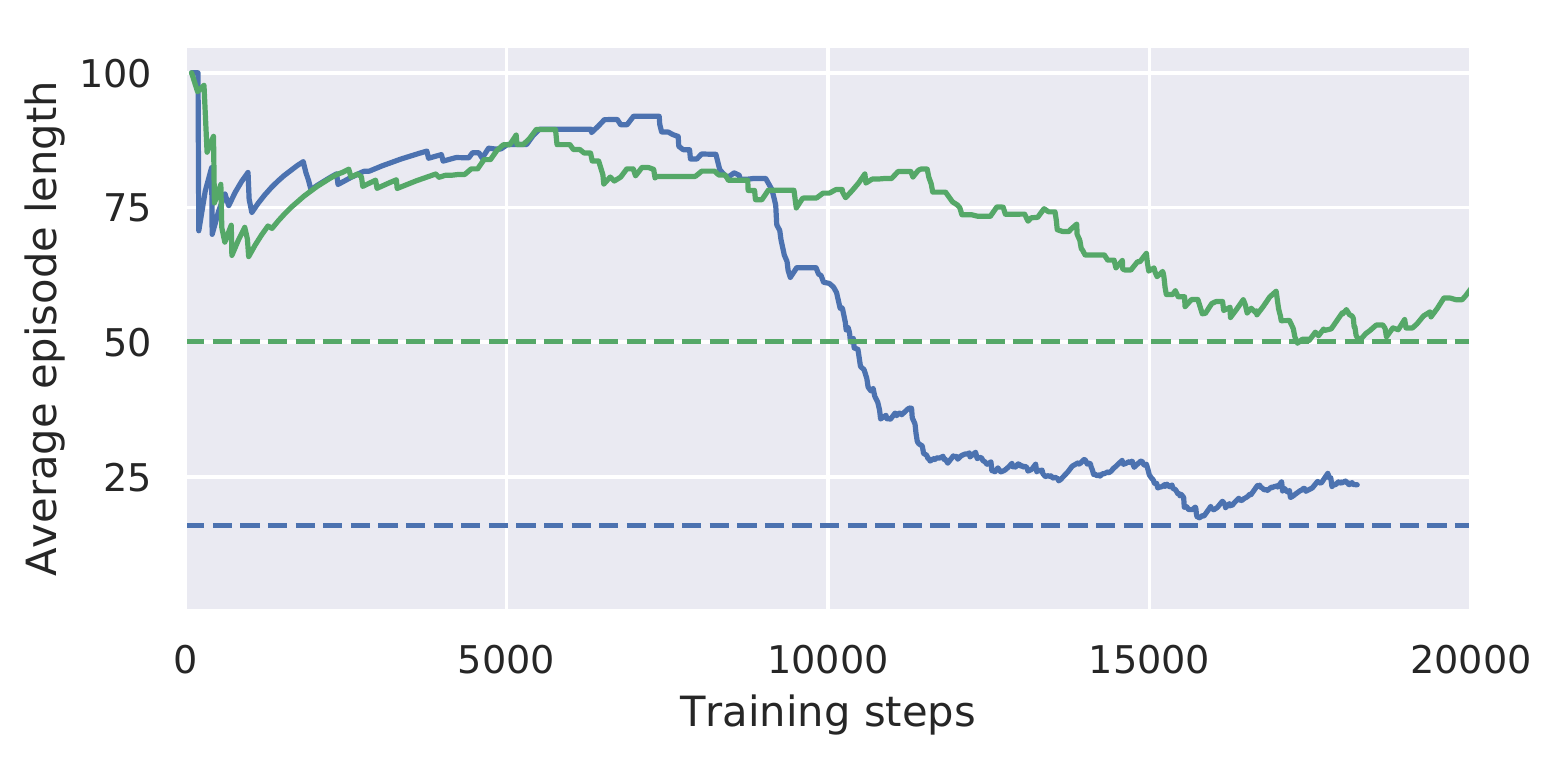}}
  \caption{Length of episode of the two Turtlebot experiment runs in different colors. Lower is better. Curves are averaged with rolling average of $50$ steps. Dashed line represents the performance of the best model obtained during training. 
  }
  \label{fig:sim-to-real experiment results}
\end{figure}

We conducted two training runs with the Turtlebot. The model of first run had success rate of $80\%$ and mean episode length $50.1$ with the best model (Figure \ref{fig:sim-to-real experiment results}). Second training run had mean success rate of $100\%$ and episode length $16.0$. Subjectively, the first model was attracted by the goal but repeatedly chose to reverse away from the goal. The second agent rotated in place until red pillar appeared to its field-of-view and began moving towards to goal, doing small fine-adjustments to stay on correct path and utilizing newly available actions appropriately.

\subsection{Experiments with removed actions}
In practical situations, changes to action space may occur from faulty or invalid hardware, preventing from executing specific actions. To study how our approach would perform in this situation, we conducted sim-to-sim transfer using DQN source model 3 with replace method with the same setup as previously. However, now one or two of the original four actions were disabled, so the agent had to find a different way to complete the task. 

\begin{figure}[h]
\centerline{\includegraphics[width=1.0\columnwidth]{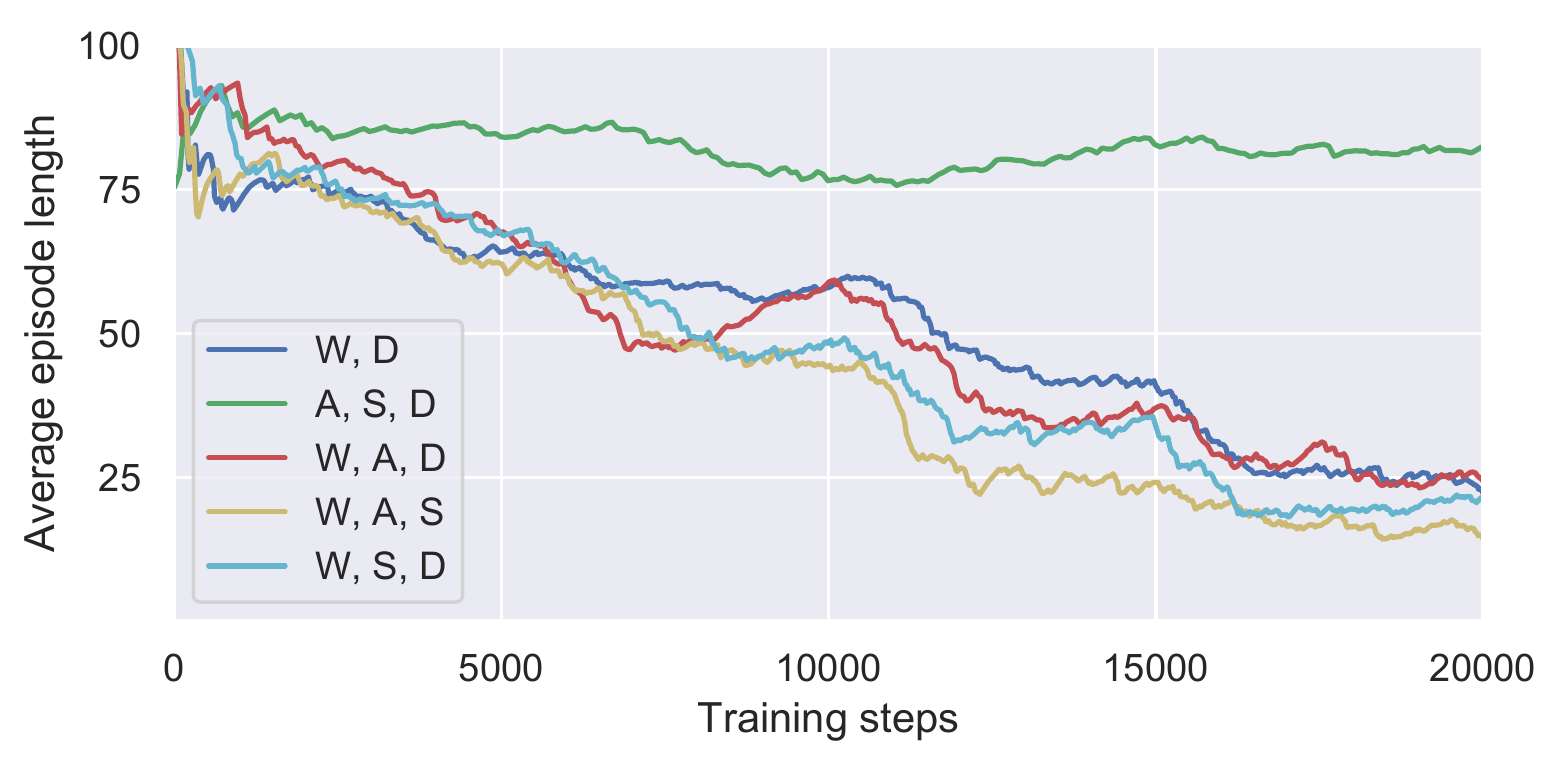}}
\caption{Agent's performance with action space where some of the previous actions are removed. Episode length is measured in time steps and the letters in labels correspond to button presses of each action (W = move forward, A = turn left, S = move backward, D = turn right). }
\label{fig:cut_out_actions}
\end{figure}

The results show that even if one of the turn actions or move backward action was removed, the agent still learned the new action space robustly in $20,000$ training steps (Figure \ref{fig:cut_out_actions}). This result suggests that with the replace method, the agent can learn the task so that it is able to adapt to action space where some of the initial actions are removed. Only when the action \say{move forward} was removed, the agent could not learn the task. This was expected, as agent does not have memory it is unable to navigate by reversing.

\section{Conclusions}

In this work we show how freezing most of the pre-trained neural network parameters can be used to effectively transfer a policy from a video game to a robot, despite the differences in action spaces between these two environments. We trained a policy on raw image data to solve a simple navigation task in Doom video game, and then successfully transferred it to a robot with a different action space where it was able to complete the same task with relatively little amount of training, including when number of available actions was reduced. These methods have promise to utilize crude simulations like video games to train policies for robots with different physical properties. 

The future work could extend the present work in terms of learning complicated abstract task in video game and then transferring to the vastly different action space structure in the physical robot. We also plan to study if Bayesian methods could be used to find good priors for the network parameters, to further speed up the learning process of new action space.

\bibliographystyle{IEEEbib}
\bibliography{strings,refs}

\end{document}